%% file: conference_041818.tex
\documentclass[conference]{IEEEtran}
\IEEEoverridecommandlockouts
\usepackage{cite}
\usepackage{url}
\usepackage{flushend}
\usepackage{amsmath,amssymb,amsfonts}
\usepackage{algorithmic}
\usepackage{graphicx}
\usepackage{textcomp}
\usepackage[table,xcdraw]{xcolor}
\usepackage{multirow}
\usepackage{caption}
\def\BibTeX{{\rm B\kern-.05em{\sc i\kern-.025em b}\kern-.08em
    T\kern-.1667em\lower.7ex\hbox{E}\kern-.125emX}}
\begin{document}

\title{Analysing Statistical methods for Automatic Detection of Image Forgery\\
}
\author{\IEEEauthorblockN{Umar Masud} \textit{\scriptsize um71000@gmail.com} 
\and
\IEEEauthorblockN{Anupam Agarwal} \textit{\scriptsize anupam@iiita.ac.in}

}

\maketitle

\begin{abstract}
Image manipulation and forgery detection have been a topic of research for more than a decade now. New-age tools and large-scale social platforms have given space for manipulated media to thrive. These media can be potentially dangerous and thus innumerable methods have been designed and tested to prove their robustness in detecting forgery. However, the results reported by state-of-the-art systems indicate that supervised approaches achieve almost perfect performance but only with particular datasets. In this work, we analyze the issue of out-of-distribution generalisability of the current state-of-the-art image forgery detection techniques through several experiments. Our study focuses on models that utilise handcrafted features for image forgery detection. We show that the developed methods fail to perform well on cross-dataset evaluations and in-the-wild manipulated media. As a consequence, a question is raised about the current evaluation and overestimated performance of the systems under consideration. 

\textcolor{red}{Note: This work was done during summer research internship at ITMR Lab, IIIT-Allahabad under supervision of Prof. Anupam Agarwal.}

\end{abstract}


\section{Introduction}
With over 4.5 billion active internet users\footnote{https://www.statista.com/statistics/617136/digital-population-worldwide/}, the amount of multimedia content being shared every day has surpassed everyone’s imagination. Large-scale and pervading social media platforms, along with easily accessible smartphones, have given rise to huge visual data such as images, videos, etc. One of the perils accompanying this huge amount of data is manipulation and malicious intent to remove the authenticity of these media. 

The availability of various image-editing software and tools such as Photoshop\footnote{https://www.photoshop.com/en}, GIMP\footnote{https://www.gimp.org/}, etc., has made it possible to create forgeries with minimal effort. Today, it is quite easy to produce a manipulated media that looks indifferent to the human eyes. Various types of digital image forgeries have evolved, the major ones include copy-move, splicing, morphing, watermarking, etc. Copy-move manipulation means cutting and pasting a portion of the same image onto itself. Splicing involves cutting and pasting from different sources. Copy-move and Splicing an image are a part of passive image forgery while examples of watermarking come under active image forgery. From an image forensics point of view, passive forgeries are of relative interest to us and hence, much research has been conducted to detect these passive image forgeries. 

Techniques designed to detect these forgeries can be divided into two classes based on how they use an image to extract its feature, one that uses handcrafted features and the other deep-learning-based extracted features. These techniques either solve for detection of forged images or detection as well as localisation of the tampered region. However, an important question of generalisability and real-world use case is found to be missing in almost every proposed method. The state-of-the-art methods have claimed the performance of classifying pristine vs tampered images with nearly 100\% accuracy. On observation, it can be seen that these near-perfect performances might be highly overestimated and are biased because of the choice of the dataset. 

In our study, we have evaluated the robustness and generalisability of five state-of-the-art methods, that utilise handcrafted features for the classification of authentic and tampered images, with a focus on copy-move and splicing forgeries. We perform a suite of evaluations by testing on a) more available datasets, both large and small b) out-of-distribution data c) amalgamating different datasets d) real-world manipulated data. 

Our results show that these near-perfect methods fail to generalise to different data and thus cannot be utilised in real-world applications. We also put forward the need to focus on cross-dataset generalisation and usage of the latest benchmark datasets, keeping in mind the real-world data as well,  to develop universal techniques for robust image forgery detection. 

\section{Related Work}

\textbf{Overview of Image Forgery Techniques:} In literature, several methods have been used to identify copy-moved and spliced images from the pristine ones. Wei et al. \cite{5652660} modelled the edge image of the image chroma component as a finite-state Markov chain feature and used Support Vector Machine (SVM) for classification. Ahmet et al. \cite{5414611} used artifacts created by Color Filter Array (CFA) processing to detect tampering. The method by Xunyu et al. \cite{10.1145/2037252.2037256} detects the noise variance differences between original and tampered parts of an image and uses its variance to classify as well as localise the tampered images. Can et al. \cite{8099686} used camera response function analysis to detect spiced images. Ankit et al. \cite{Jaiswal2020} used a hybrid feature set features extracted from techniques comprising of Laws Texture Energy (LTE), Local Binary Pattern (LBP), Histogram of Oriented Gradients (HoG), Discrete Wavelet Transform (DWT) to make a classification using SVM. Habibi et al. \cite{habibi} proposed a blind image splicing detection method based on the colour distribution in the neighbourhood of edge pixels. An enhanced technique used by Navdeep et al. \cite{Kanwal2020} was to use the optimal threshold-based local ternary patterns to classify images.  Moghaddasi et al. \cite{Moghaddasi2019} had used Singular Value Decomposition on Discrete Cosine Transform (DCT) coefficients to extract the features. Abrahim et al. \cite{Abrahim2019} used Local Binary Pattern (LBP), Higher-order statistical features, and HoG to finally train an artificial neural network for classification.

\input{tables/t_dataset_Description}

Several deep-learning approaches have also been used to detect/localise tampered images.  Fatima et al. \cite{ElBiach2021} used a network of the U-Net \cite{ronneberger2015unet} family as encoder/decoder to create binary masks of tampered regions. Rao et al. \cite{7823911} used a specially designed CNN to train labelled patches of images to extract feature-set followed by SVM for classification. Salloum et al. \cite{SALLOUM2018201} employed a Multi-Task Fully Convolutional Network to learn surface labels as well as edge or boundary labels of the spliced region.  Yue et al. \cite{8953774} introduced a novel end-to-end Deep Neural Network solution to image forgery localization called ManTra-Net. 
Some research methods have also used GANs \cite{goodfellow2014generative} for solving image forgery detection. Methods by Younis et al. \cite{info10090286}, Liu et al. \cite{8658131}, Ashraful et al. \cite{9157762}, Vladimir et al. \cite{NEURIPS2019_98dce83d}, have made use of GANs in different ways to detect various image manipulations.

A more extensive overview for image forgery methods has been covered by these survey papers Navdeep et al. \cite{Kaur_Kanwal_2019}, Zheng et al. \cite{ZHENG2019380}, Sabeena et al. \cite{abraham2020digital}, and Abhishek et al. \cite{Kashyap2017AnEO}.

\textbf{Methods Analysed in this Study:} In our study, we have chosen five methods \cite{Alahmadi2017}, \cite{DUA2020369}, \cite{9349611},  \cite{10.1007/978-981-10-2738-3_27}, and \cite{electronics9091500}, that are based on statistical approaches and use hand-crafted features and white-box models (classical machine learning algorithms) to classify pristine and tampered images. Alahmadi et al. \cite{Alahmadi2017} used textural pattern LBP and frequency domain DCT to extract features for classification using SVM. The method achieved 97.00\% on CASIA v1.0, 97.50\% on CASIA v2.0 and 97.77\% on the Columbia Colored dataset. Shilpa et al. \cite{DUA2020369} made use of artifacts that originated due to manipulations of JPEG encoded images, extracting useful features through standard deviation and number of ones in the AC components of DCT coefficients. The achieved average detection rates for CASIA v1.0 and CASIA v2.0 are above 93\% and 98\% respectively. Arman et al. \cite{9349611} proposed a method that uses extracted features from the mantissa distribution of DCT coefficients. The reported results are 99.78\% on the CASIA v1.0 dataset. Mandeep et al. \cite{10.1007/978-981-10-2738-3_27} used Discrete Wavelet Transform and histogram of Local Binary Patterns to extract features for classification. The best results obtained were up to 97\% on JPEG images in the spliced dataset.
Mohammad et al. \cite{electronics9091500} proposed a method based on Discrete Cosine Transformation (DCT) and Local Binary Pattern (LBP) and a new feature extraction method using the mean summary statistic. The best results reported were 99.55\%, 99.88\%, and 98.20\% on CASIA v1.0, CASIA v2.0, and Columbia Color datasets respectively. Details of replication of these methods are provided in Table \ref{tab:reproduced}.

While there are plenty of methods to choose from, we selected the above five methods because a) statistical approaches using white-box (classical machine learning) models, thus having explainability b) used common datasets c) proposed nearly perfect accuracies d) reproducible and computationally efficient to run (check section \ref{section:experiments}).

\textbf{Nature of Study and its relevance:} Across the field of machine learning and its applications, we come across methods that report high accuracy on the datasets that the models are trained on, but fail to perform well on test datasets that are slightly different in distribution. In the case of Image Forgery, K.Asghar et al. \cite{Asghar2019} tested the efficacy of their proposed method by training and testing on the same datasets, as well as did a comprehensive cross dataset evaluation. They talked about the generalisability of their method through testing on out-of-distribution data, a mixture of datasets, etc.. Isaac et al. \cite{10.1007/s11042-017-5189-5} also showed a cross-dataset evaluation of their proposed method in one of their ablations. Even outside the field of image forgery, there have been growing interest in out of distribution and cross dataset evaluations for computer vision and natural language processing. For example, in the area of hate speech detection, Arango et.al \cite{10.1145/3331184.3331262} discussed how the state-of-art hate speech models, fail to perform in the wild. Inspired by this body of work, we took up the study of analysing the state-of-art statistical methods for image forgery detection and classification, performing a series of experiments to comment on the robustness of the techniques.

\section{Datasets Description}
\label{section:datasets}
The most common and popularly used image forgery datasets that are publicly available for research are CASIA v1.0 \cite{6625374}, CASIA v2.0 \cite{6625374}, MICC-F220 \cite{5734842}, MICC-F2000 \cite{5734842}, Columbia Colour \cite{4036658}, COVERAGE \cite{7532339}, and CG-1050 v2 \cite{CASTRO2020104864}. More recently, some large-scale benchmark datasets have been released to facilitate the work in image forensics like DEFACTO \cite{8903181} and MFC \cite{518026}, \cite{518286}  datasets. Additionally, the advent of the Internet has increased the number of forged images generated and circulated. To keep up with new trends in images across the Internet, some in-the-wild datasets have been proposed. These are useful in verifying the robustness of forgery detection techniques; datasets such as WildWeb \cite{7169839}, PS-Battles \cite{heller2018psbattles}, IMD-2020 Real Life Manipulated Images \cite{9096940}, and FRITH \cite{Asghar2019}. A detailed description of each of these datasets is mentioned in Table \ref{tab:datasetoverview}.  In our study, we have utilised these datasets under different settings, details of which are discussed in Section \ref{section:experiments}.

To make better evaluations and corroborate our study, we have divided the datasets into two categories based on their creation:
\begin{itemize}
    \item \textbf{Controlled-environment Datasets:} Datasets such as CASIA v1.0 \cite{6625374}, CASIA v2.0 \cite{6625374}, MICC-F220 \cite{5734842}, MICC-F2000 \cite{5734842}, Columbia Color \cite{4036658}, COVERAGE \cite{7532339}, CG-1050 v2 \cite{CASTRO2020104864}, DEFACTO \cite{8903181} and MFC \cite{518026}, \cite{518286}. These are the commonly used standard datasets, annotated by academicians of the community to serve the purpose in image forensic tasks. To refine our study, we have further divided these into sub-classes based on their relative size:
    \begin{itemize}
        \item \textbf{Large Scale Datasets:} Having $\geq1400$ images.
        \item \textbf{Medium Scale Datasets:} Having 400-1400 images.
        \item \textbf{Small Scale Datasets:} Having $\leq400$ images.
    \end{itemize}
    However, we do not use datasets of medium-scale in our study. 
    Also, another sub-division is made based on of the time of the release of these datasets: 
    \begin{itemize}
        \item \textbf{Relatively Older Datasets:} That were released before 2016, made using older photo-editing tools.
        \item \textbf{Relatively Newer Datasets:} That were released after 2016, and have been made using relatively newer photoshopping tools and software.
    \end{itemize}
    \item \textbf{In-the-Wild Datasets:} such as WildWeb \cite{7169839}, PS-Battles \cite{heller2018psbattles}, IMD-2020 Real Life Manipulated Images \cite{9096940}, and FRITH \cite{Asghar2019} in which the images are collected from an uncontrolled environment. It contains examples of real-life image forgeries taken from the internet and elsewhere.
\end{itemize}

\section{Overview of the Analysis}
Out of our selected papers \cite{Alahmadi2017}, \cite{DUA2020369}, \cite{9349611},  \cite{10.1007/978-981-10-2738-3_27}, and \cite{electronics9091500} all have either explicitly mentioned that they are solving for both copy-move and splice forgeries or they have used a dataset that has both copy-move and spliced images, thus allowing us to assume that they have considered both type of forgeries.

We have divided our experiments into 3 sets, each set itself has an array of experiments spanning a combination of multiple datasets and the methods discussed in Section \ref{section:experiments}. 
\begin{itemize}
    \item SET 1: Different experiments of training/testing on the same data.
    \item SET 2: Cross-dataset evaluations.
    \item SET 3: Experiments with in-the-wild data.
\end{itemize}

\section{Experimental Results, Comparison and Discussion}
\label{section:experiments}

\textbf{Experimental Setup:} All the experiments were run using Intel  i5-5300U CPU @ 2.30GHz × 4, 12Gb RAM, Ubuntu 18.04.5 LTS, and Python 3.6.9 environment.

\textbf{Evaluation Criteria:} In all our results, the evaluation metric that we have mentioned are accuracy percentage and macro-F1 score (in brackets), since it is a binary classification problem with mostly balanced classes. Places marked with (*) in Tables \ref{tab:olddata}, \ref{tab:newage}, \ref{tab:largetrain}, \ref{tab:smalltrain}, \ref{tab:realtest} signifies that in that result, either the authentic or the tampered class were not identified at all and that class' precision = 0 and recall = 0.

\textbf{Hyperparameter Tuning and Classification:} We have used SVM as a classifier with the same kernel (RBF or poly) as in the original papers for each. The hyperparameters are tuned through GridSearchCV with 10 K-Fold splits for all the results to get the highest performance. 

Some noticeable observations are discussed in the following subsections.

\input{tables/t_reproduced}

\subsection{\textbf{SET 1}}

\subsubsection{\textbf{Reproduced Results}}
In Table \ref{tab:reproduced}, we have shown the replication results of our 5 methods on the datasets used by the respective papers. The accuracy percentage we got is mentioned along with the proposed ones (in brackets). Since no source code was available for any of the methods, we had to implement each of the proposed methods from scratch\footnote{All the reproduced code will be released in public}. While we achieved nearly similar accuracies for all, in two of the methods, namely Alahmadi et al. \cite{Alahmadi2017} and Mohammad et al. \cite{electronics9091500}, we found the accuracies to be significantly less than the reported ones for Columbia Color and Columbia Gray datasets. However, since the results on other datasets of the paper were nearly the same for these two methods, we concluded that the replication was right and fairly done.

\input{tables/t_Extended_OldData}

\input{tables/t_Extended_NewAge}
\input{tables/t_MixtureCommonDatasets}
\input{tables/t_LargeTrainRestTest}

\input{tables/t_SmallTrainRestTest}
\subsubsection{\textbf{Extending training/testing on other common Image Forgery Datasets}}
From the major standard datasets discussed in Section \ref{section:datasets}, we find that the papers we reviewed did not test their systems on all of the standard datasets. For example, while \cite{DUA2020369} reported performances on CASIA v1.0 \cite{6625374} and CASIA v2.0 \cite{6625374} datasets, they did not report any results for other datasets such Columbia Color \cite{4036658}, MICC-F220 \cite{5734842}, etc. Meanwhile, the proposed method by \cite{Alahmadi2017} reported a high performance across datasets CASIA v1.0 \cite{6625374}, CASIA v2.0 \cite{6625374}, and Columbia Color \cite{4036658} but did not review the datasets COVERAGE \cite{7532339}, MICC-F220 \cite{5734842}, etc. So, as the first step towards better understanding the robustness of different methods, we made sure all proposed methods are tested on other standard datasets as well. For these experiments, we trained and tested on the same datasets. The results for these experiments are shown in Table \ref{tab:olddata}.

It can be seen that there is a significant performance drop. For the COVERAGE \cite{7532339} dataset, of size 200, it was observed that the performance of methods went below 50\% in both accuracy and F1 score. It means that the methods were unable to generate distinguishable features at all. While for an equally small dataset of size 220, the MICC-F220 \cite{5734842} reported far better results. Thus, showing that the number of images did not affect the performance in this case.
\input{tables/t_CommonTrainRealWorldTest}

\subsubsection{\textbf{Extending training/testing on new-age, less common Image Forgery Datasets}}
In the previous subsection, we extended our methods on the commonly used datasets that are relatively older. In this setting, we have used three relatively newer datasets - CG-1050 v2 \cite{CASTRO2020104864}, DEFACTO \cite{8903181} and MFC \cite{518026}, \cite{518286}, to test the performance of the proposed methods. With advancements, the tools and software for photo-editing have also evolved and thus newer, sharper, and more difficult forgeries can be created today. This need for future-proofing and long-term robustness should be kept in mind while we are developing methods to detect forgeries. This is the reason why we differentiated the datasets into relatively old and new. The results are shown in Table \ref{tab:newage}.

It can be seen that accuracy has dropped overall. DEFACTO \cite{8903181} dataset, although having 10,000 image samples, performed poorly in comparison to MFC \cite{518026}\cite{518286}  dataset which had only around 1200 samples. CG-1050 v2 \cite{CASTRO2020104864} had around 2100 samples but still performed the worst. Again showing that the number of images in the dataset does not guarantee better performance.

\subsubsection{\textbf{Amalgamation of commonly used datasets}}
\label{sssec:mixcommon}
In this evaluation, we made a mixture of different datasets to increase the variety of our evaluation samples and trained/tested the methods on the same. We took 100 pristine and 100 tampered samples randomly from the datasets CASIA v1.0 \cite{6625374}, CASIA v2.0 \cite{6625374}, MICC-F220 \cite{5734842}, MICC-F2000 \cite{5734842}, COVERAGE \cite{7532339}, and Columbia Color \cite{4036658}, and performed the whole evaluation on a set of 3 random selections. Three random selections were made to remove any bias of choice. Finally, we reported the mean of the results obtained from the three random selections. The results are shown in Table \ref{tab:mixcommon}. 

From the results, we see that although the samples from 4 out of our 6 datasets used in the mixture were from high performing datasets as observed in the previous experiments, the overall performance of the amalgamated dataset skewed to the lower side. Thus, it can be said that samples from datasets, namely COVERAGE \cite{7532339} and Columbia Color \cite{4036658}, were low performing (as seen in previous experiments), influenced the final results more. For the future, we can say that while creating a mixed dataset for evaluation purposes we should take care to balance the data samples so that the models do not underfit or overfit, and generalise well.

\subsection{\textbf{SET 2}}
 While in the first set, we took the train-test split from the same dataset, we also wanted to explore the cross-dataset performance of the methods, as a means of understanding their predictions on out-of-distribution samples. This approach has been divided into two parts, in the first part we train on relatively large scale datasets like CASIA v1.0 \cite{6625374}, CASIA v2.0 \cite{6625374} and MICC-F2000 \cite{5734842} which have around $\geq1400$ images. In the second part, we train on relatively small scale datasets like MICC-F220 \cite{5734842}, COVERAGE \cite{7532339} and Columbia Color \cite{4036658} which have $\leq400$ images. 

We have not used the CG-1050 v2 \cite{CASTRO2020104864}, DEFACTO \cite{8903181} and MFC \cite{518026}, \cite{518286}  datasets in cross-evaluation as they are relatively newer datasets. So we restricted to the standardly used old datasets only on which most of the research has been conducted in the past several years.

The idea behind this division is to capture the out-of-distribution (OOD) tests as well as low-resource availability and see how these two factors affect the performance of forgery detection. 
In both scenarios, a method trained on one standard dataset (either large-scale or small-scale) is tested on all available standard ones. This led to a combination of 15 train-test sets per paper. The results are shown in Table \ref{tab:largetrain} and \ref{tab:smalltrain}.

It can be seen that in general, all the methods have performed poorly on all cross-dataset evaluations. Even the datasets on which the training accuracies were fairly good, the cross-tests have given below-par results. For the case of a large scale dataset, we took CASIA v1.0 \cite{6625374} and found the average difference in the drop in performance in training and when tested on OOD samples across other datasets, as well as across the five methods. The average drop was found to be around 40.6\%. We followed the same procedure for a low-scale dataset MICC-F220 \cite{5734842} which had given fairly good training accuracies, the average drop in performance was about 38.6\%. 

\subsection{\textbf{SET 3}}

\subsubsection{\textbf{Training on Commonly used Image Forgery Datasets and Testing on	Real-World Manipulated Data}}
In the previous sections of our experiments, we only considered the controlled environment datasets, to test the completeness and robustness of our methods. However, the ultimate aim of image forgery systems is to be deployed in the real world, where we can detect and flag such forged images in real-time, as these in-the-wild images can differ in terms of quality and quantity and are not generated (or annotated) by expert users. Their dataset distribution can vary from the datasets we have covered so far. Hence to further the usability and robustness of our chosen methods, we train the methods on the standard datasets and then test them on various in-the-wild datasets- WildWeb \cite{7169839}, PS-Battles \cite{heller2018psbattles}, IMD-2020 Real Life Manipulated Images \cite{9096940}, and FRITH \cite{Asghar2019}. This simulates how the methods trained in a controlled environment would perform in the real world.
Table \ref{tab:realtest} shows the results obtained.

The results show that the methods trained on the commonly used datasets could not give a good performance on the in-the-wild data. The best that any performance could reach was just 74.40\% by  Alahmadi et al. \cite{Alahmadi2017} in IMD-2020 Real Life Manipulated Images \cite{9096940} test data, trained on CASIA v2.0 \cite{6625374} dataset. On average, the performance on the test datasets has been better when the training datasets were CASIA v1.0 \cite{6625374} and CASIA v2.0 \cite{6625374}. The worst performance is when the training dataset is MICC-F2000 \cite{5734842}. Again, the size of the training datasets played no significant role in determining the performances on the test datasets.

\subsubsection{\textbf{Amalgamation of in-the-wild datasets}}
Like in Section \ref{sssec:mixcommon}, here also we have made a mixed dataset using the in-the-wild datasets - WildWeb \cite{7169839}, PS-Battles \cite{heller2018psbattles}, IMD-2020 Real Life Manipulated Images \cite{9096940}, and FRITH \cite{Asghar2019}. We took 150 pristine and 150 tampered samples randomly from each of the datasets. This gives our dataset variety by having different types of images and various forgeries. We trained and tested the methods on this same data for three random selections. Finally, a mean performance is noted. The purpose for testing on this mixture of datasets is to check if the methods can work better for real-life use cases if they are allowed to be trained and tested on in-the-wild data which comes from actual image forgery scenarios. The results are shown in Tabe \ref{tab:mixreal}.
\input{tables/t_MixtureRealDatasets}

As seen from the results, the methods failed to give a satisfactory performance. The best result was just 63.61\% by Alahmadi et al. \cite{Alahmadi2017}. It can also be seen that the methods perform worse on a mixture of in-the-wild datasets as against a mixture of controlled environment datasets, seen in Section \ref{sssec:mixcommon}. This raises a big doubt on the real-life usability of these methods and forces us to think towards developing more comprehensive techniques aimed at the purpose of solving image forgery in the practical world rather than just out-performing the existing state-of-the-art.

\section{Conclusion}
Through our experiments, we have shown that the methods for image forgery detection are far from being robust in the true sense. Their performance relies heavily on the choice of the datasets and does not work well across different evaluations. For successful practical applications, we should look for techniques that can generalise to unseen data. Work should be done keeping in mind the universal need and application of the methods.

The future scope of this paper is to perform cross-evaluation studies on methods other than the statistical, white-box model ones. Deep-learning based techniques should also be explored for generalisability and real-world use cases. Another piece of work can be done to extend this study on techniques of localisation of forgery as well.

\bibliographystyle{./bibliography/IEEEtran}
\bibliography{./bibliography/citations}
\end{document}

%% file: tables/t_dataset_Description.tex
\begingroup
\setlength{\tabcolsep}{6pt} 
\renewcommand{\arraystretch}{1.2}
\begin{table*}
\centering
\resizebox{\textwidth}{!}{%
\begin{tabular}{|c|c|c|c|c|c|c|c|}
\hline
\textbf{Dataset} &
  \textbf{Release year} &
  \textbf{Forgery type} &
  \textbf{Pristine/Tampered} &
  \textbf{\begin{tabular}[c]{@{}c@{}}Image \\ Size\end{tabular}} &
  \textbf{Post-Processing} &
  \textbf{\begin{tabular}[c]{@{}c@{}}In-the-Wild\\ Data\end{tabular}} &
  \textbf{\begin{tabular}[c]{@{}c@{}}Sub-Sample\\ Used\end{tabular}} \\ \hline
Columbia Color    & 2006 & Splice                                                                    & 183/180     & Various   & No  & No  & 183/180   \\ \hline

MICC-F220         & 2011 & Copy-Move                                                                 & 110/110     & Various   & No  & No  & 110/110   \\ \hline

MICC-F2000        & 2011 & Copy-Move                                                                 & 1300/700    & 2048x1536 & No  & No  & 700/700   \\ \hline

CASIA v1.0        & 2013 & \begin{tabular}[c]{@{}c@{}}Splicing,\\ Copy-Move\end{tabular}                                                                    & 800/921     & 384x256   & No  & No  & 800/921   \\ \hline
CASIA v2.0        & 2013 & \begin{tabular}[c]{@{}c@{}}Splice,\\ Copy-Move\end{tabular}               & 7491/5123   & 384x256   & Yes & No  & 7491/5123 \\ \hline

Wild Web          & 2015 & \begin{tabular}[c]{@{}c@{}}Splice,\\ Copy-Move,\\ Erase-fill\end{tabular} & 90/13577    & Various   & Yes & Yes & 90/99     \\ \hline

COVERAGE          & 2016 & Copy-Move                                                                 & 100/100     & Various   & No  & No  & 100/100   \\ \hline

PS-Battles        & 2018 & \begin{tabular}[c]{@{}c@{}}Splicing,\\ Copy-Move\end{tabular}             & 11142/91886 & Various   & Yes & Yes & 498/500   \\ \hline

MFC18 dev-1 ver-1 & 2018 & Splicing                                                                  & 997/5711    & Various   & No  & No  & 614/616   \\ \hline

FRITH             & 2019 & \begin{tabular}[c]{@{}c@{}}Splicing,\\ Copy-Move\end{tabular}             & 156/229     & Various   & Yes & Yes & 156/229   \\ \hline

CG-1050 v2        & 2019 & \begin{tabular}[c]{@{}c@{}}Splicing,\\ Copy-Move\end{tabular}             & 1050/1050   & Various   & Yes & No  & 1050/1050 \\ \hline

DEFACTO &
  2019 &
  \begin{tabular}[c]{@{}c@{}}Splicing,\\ Copy-Move\end{tabular} &
  \begin{tabular}[c]{@{}c@{}}105000 spliced,\\ 19000 copy-moved\end{tabular} &
  Various &
  No &
  No &
  5000/5000 \\ \hline

\begin{tabular}[c]{@{}c@{}}IMD2020 Real-Life \\ Manipulated Images\end{tabular} &
  2020 &
  \begin{tabular}[c]{@{}c@{}}Splice, \\ Copy-Move, \\ Re-touching\end{tabular} &
  414/2010 &
  Various &
  Yes &
  Yes &
  414/419 \\ \hline
  
\end{tabular}%
}
\caption{Description of Datasets and sub-samples used} 
\label{tab:datasetoverview}
\end{table*}
\endgroup

%% file: tables/t_reproduced.tex
\begin{table}
\renewcommand{\arraystretch}{1.8}
\resizebox{\columnwidth}{!}{%
\begin{tabular}{|c|c|c|c|c|}
\hline
\multirow{2}{*}{\textbf{Methods}} & \multicolumn{4}{c|}{\textbf{Datasets}}                                    \\ \cline{2-5} 
 & \textit{\textbf{CASIA v1.0}} & \textit{\textbf{CASIA v2.0}} & \textit{\textbf{Columbia Color}} & \textit{\textbf{Columbia Gray}} \\ \hline
Alahmadi et al. \cite{Alahmadi2017}          & 96.80\%  (97.00) & 97.10\%  (97.50) & 80.02\%  (97.77) & -                \\ \hline
Shilpa et al.  \cite{DUA2020369}            & 96.81\%  (93.20) & 97.34\%  (98.30) & -                & -                \\ \hline
Arman et al. \cite{9349611}             & 95.50\%  (99.80) & -                & -                & -                \\ \hline
Mandeep et al. \cite{10.1007/978-981-10-2738-3_27}           & 93.67\%  (92.62) & 97.14\%  (97.34) & 87.2\%  (87.05)  & 74.44\%  (75.93) \\ \hline
Mohammed et al. \cite{electronics9091500}          & 95.65\%  (99.50) & 96.86\%  (99.88) & 72.60\%  (98.20) & 75.55\%  (85.56) \\ \hline
\end{tabular}%
}
\caption{Reproduced Results}
\label{tab:reproduced}
\end{table}

%% file: tables/t_Extended_OldData.tex
\begingroup
\begin{table}
\renewcommand{\arraystretch}{1.5}
\centering
\resizebox{\columnwidth}{!}{%
\begin{tabular}{| 
>{\columncolor[HTML]{FFFFFF}}c |
>{\columncolor[HTML]{FFFFFF}}c |
>{\columncolor[HTML]{FFFFFF}}c |
>{\columncolor[HTML]{FFFFFF}}c |}
\hline
\cellcolor[HTML]{FFFFFF}                                   & \multicolumn{3}{c|}{\cellcolor[HTML]{FFFFFF}\textbf{Datasets}}                
\\ \cline{2-4} 
\multirow{-2}{*}{\cellcolor[HTML]{FFFFFF}\textbf{Methods}} & \textit{\textbf{MICC-220}} & \textit{\textbf{MICC-2000}} & \textit{\textbf{COVERAGE}}  \\ \hline
Alahmadi et al. \cite{Alahmadi2017}                                            & 90.90\% (0.91)             & 95.35\% (0.95)              & 47.50\% (0.32)*                                      \\ \hline
Shilpa et al. \cite{DUA2020369}                                             & 93.18\% (0.93)             & 97.5\% (0.97)               & 42.5\% (0.42)                                         \\ \hline
Arman et al. \cite{9349611}                                              & 84.84\% (0.84)             & 93.33\% (0.93)              & 46.67\% (0.32)*                                       \\ \hline
Mandeep et al. \cite{10.1007/978-981-10-2738-3_27}                                             & 93.18\% (0.93)             & 97.85\% (0.98)              & 47.5\% (0.32)                                         \\ \hline
Mohammed et al. \cite{electronics9091500}                                            & 88.63\% (0.87)             & 88.92\% (0.89)              & 42.5\% (0.30)                                         \\ \hline
\end{tabular}
}
\caption{Extended to old datasets}
\label{tab:olddata}
\end{table}
\endgroup

%% file: tables/t_Extended_NewAge.tex
\begingroup
\begin{table}
\renewcommand{\arraystretch}{1.5}
\centering
\resizebox{\columnwidth}{!}{%
\begin{tabular}{|c|c|c|c|}
\hline
\multirow{2}{*}{\textbf{Methods}} & \multicolumn{3}{c|}{\textbf{Datasets}}                                             \\ \cline{2-4} 
                                  & \textit{\textbf{MFC18}} & \textit{\textbf{DEFACTO}} & \textit{\textbf{CG-1050 v2}} \\ \hline
Alahmadi et al. \cite{Alahmadi2017}                  & 90.65\%  (0.91)         & 74.81\%  (0.73)           & 45.20\% (0.31)*              \\ \hline
Shilpa et al. \cite{DUA2020369}                    & 87.80\%  (0.88)         & 75.71\%  (0.75)           & 58.90\% (0.59)               \\ \hline
Arman et al. \cite{9349611}                   & 84.14\%  (0.84)         & 73.47\%  (0.73)           & 61.64\% (0.61)               \\ \hline
Mandeep et al. \cite{10.1007/978-981-10-2738-3_27}                  & 91.05\%  (0.91)         & 73.31\%  (0.72)           & 45.20\% (0.31)               \\ \hline
Mohammed et al. \cite{electronics9091500}                   & 85.77\%  (0.86)         & 74.62\%  (0.73)           & 49.65\% (0.36)               \\ \hline
\end{tabular}%
}
\caption{Extended to new-age, latest datasets}
\label{tab:newage}
\end{table}
\endgroup

%% file: tables/t_MixtureCommonDatasets.tex
\begingroup
\renewcommand{\arraystretch}{1.7}
\begin{table}
\centering
\resizebox{\columnwidth}{!}{%
\begin{tabular}{ccccc}
\multirow{2}{*}{\textbf{Methods}} & \multicolumn{4}{c}{\textbf{Data Samples}}                             \\ \cline{2-5} 
 & \textit{\textbf{Random Sample 1}} & \textit{\textbf{Random Sample 2}} & \textit{\textbf{Random Sample 3}} & \textit{\textbf{Mean Performance}} \\ \hline
Alahmadi et al. \cite{Alahmadi2017}                   & 73.75\%  (0.74) & 71.25\%  (0.71) & 74.58\%  (0.75) & 73.19\%  (0.73) \\ \cline{1-1}
Shilpa et al. \cite{DUA2020369}                    & 75.00\%  (0.75) & 74.58\%  (0.75) & 78.33\%  (0.78) & 75.97\%  (0.76) \\ \cline{1-1}
Arman et al. \cite{9349611}                      & 67.08\%  (0.67) & 74.58\%  (0.75) & 76.66\%  (0.77) & 72.77\%  (0.73) \\ \cline{1-1}
Mandeep et al. \cite{10.1007/978-981-10-2738-3_27}                   & 72.08\%  (0.72) & 68.75\%  (0.69) & 72.08\%  (0.72) & 70.97\%  (0.71) \\ \cline{1-1}
Mohammed et al. \cite{electronics9091500}                   & 66.25\%  (0.66) & 66.67\%  (0.66) & 67.94\%  (0.67) & 66.94\%  (0.67)
\end{tabular}%
}
\caption{Amalgamation of Common Datasets}
\label{tab:mixcommon}
\end{table}
\endgroup

%% file: tables/t_LargeTrainRestTest.tex
\begingroup
\setlength{\tabcolsep}{10pt} 
\renewcommand{\arraystretch}{1.5}
\begin{table*}[ht!]
\centering
\resizebox{\textwidth}{!}{%
\begin{tabular}{cccccccc}
\multirow{2}{*}{\textbf{Method}} &
  \multirow{2}{*}{\textbf{\begin{tabular}[c]{@{}c@{}}Train\\ Dataset\end{tabular}}} &
  \multicolumn{6}{c}{\textbf{Test Datasets}} \\ \cline{3-8} 
 &
   &
  \textit{\textbf{CASIA 1.0}} &
  \textit{\textbf{CASIA 2.0}} &
  \textit{\textbf{MICC-220}} &
  \textit{\textbf{MICC-2000}} &
  \textit{\textbf{COVERAGE}} &
  \textit{\textbf{Columbia Color}} \\ \hline
\multirow{3}{*}{Alahmadi et al. \cite{Alahmadi2017}} &
  \textit{\textbf{CASIA 1.0}} &
  - &
  73.63\%  (0.69) &
  42.42\%  (0.42) &
  65.95\%  (0.66) &
  46.66\%  (0.32)* &
  52.29\%  (0.34) \\ \cline{2-2}
 &
  \textit{\textbf{CASIA 2.0}} &
  61.12\%  (0.61) &
  - &
  57.57\%  (0.56) &
  55.00\%  (0.51) &
  45.00\%  (0.33) &
  63.30\%  (0.59) \\ \cline{2-2}
 &
  \textit{\textbf{MICC-2000}} &
  51.25\%  (0.34)* &
  60.73\%  (0.38) &
  46.96\%  (0.32)* &
  - &
  53.30\%  (0.35)* &
  46.70\%  (0.32)* \\ \cline{1-2}
\multirow{3}{*}{Shilpa et al. \cite{DUA2020369}} &
  \textit{\textbf{CASIA 1.0}} &
  - &
  79.30\%  (0.77) &
  42.42\%  (0.40) &
  49.52\%  (0.48) &
  48.33\%  (0.37) &
  58.71\% (0.58) \\ \cline{2-2}
 &
  \textit{\textbf{CASIA 2.0}} &
  66.34\% (0.64) &
  - &
  46.96\%  (0.47) &
  51.90\%  (0.51) &
  56.66\% (0.53) &
  47.70\%  (0.48) \\ \cline{2-2}
 &
  \textit{\textbf{MICC-2000}} &
  51.25\%  (0.34)* &
  60.66\%  (0.38)* &
  46.96\%  (0.32)* &
  - &
  53.30\%  (0.35)* &
  47.8\%  (0.32)* \\ \cline{1-2}
\multirow{3}{*}{Arman et al. \cite{9349611}} &
  \textit{\textbf{CASIA 1.0}} &
  - &
  39.15\%  (0.31) &
  48.48\%  (0.45) &
  46.67\%  (0.44) &
  50.00\% (0.49) &
  62.38\%  (0.62) \\ \cline{2-2}
 &
  \textit{\textbf{CASIA 2.0}} &
  67.50\% (0.63) &
  - &
  37.87\%  (0.35) &
  50.00\%  (0.46) &
  48.30\%  (0.39) &
  46.78\%  (0.46) \\ \cline{2-2}
 &
  \textit{\textbf{MICC-2000}} &
  48.74\%  (0.41) &
  60.68\%  (0.38) &
  46.96\%  (0.40) &
  - &
  53.33\%  (0.35)* &
  47.70\%  (0.34) \\ \cline{1-2}
\multirow{3}{*}{Mandeep et al. \cite{10.1007/978-981-10-2738-3_27}} &
  \textit{\textbf{CASIA 1.0}} &
  - &
  79.94\%  (0.79) &
  57.57\%  (0.50) &
  55.47\%  (0.55) &
  45.00\%  (0.42) &
  53.21\%  (0.36) \\ \cline{2-2}
 &
  \textit{\textbf{CASIA 2.0}} &
  68.08\%  (0.64) &
  - &
  54.54\%  (0.52) &
  56.90\%  (0.53) &
  46.67\%  (0.44) &
  47.70\%  (0.32)* \\ \cline{2-2}
 &
  \textit{\textbf{MICC-2000}} &
  51.25\%  (0.34)* &
  60.66\%  (0.38)* &
  46.96\%  (0.32)* &
  - &
  53.33\%  (0.35)* &
  47.60\%  (0.32)* \\ \cline{1-2}
\multirow{3}{*}{Mohammed et al. \cite{electronics9091500}} &
  \textit{\textbf{CASIA 1.0}} &
  - &
  64.78\%  (0.64) &
  59.09\%  (0.55) &
  63.09\%  (0.62) &
  46.67\%  (0.32)* &
  46.78\%  (0.42) \\ \cline{2-2}
 &
  \textit{\textbf{CASIA 2.0}} &
  56.28\%  (0.56) &
  - &
  48.48\%  (0.42) &
  49.52\%  (0.39) &
  45.00\%  (0.37) &
  42.20\%  (0.42) \\ \cline{2-2}
 &
  \textit{\textbf{MICC-2000}} &
  52.22\%  (0.42) &
  56.80\%  (0.46) &
  59.00\%  (0.56) &
  - &
  45.00\%  (0.45) &
  45.87\%  (0.45) \\ \cline{2-2}
\end{tabular}%
}
\caption{Training on Large Datasets and Testing on the rest}
\label{tab:largetrain}
\end{table*}
\endgroup

%% file: tables/t_SmallTrainRestTest.tex
\begingroup
\setlength{\tabcolsep}{10pt} 
\renewcommand{\arraystretch}{1.5}
\begin{table*}
\centering
\resizebox{\textwidth}{!}{%
\begin{tabular}{cccccccc}
\multirow{2}{*}{\textbf{Method}} &
  \multirow{2}{*}{\textbf{\begin{tabular}[c]{@{}c@{}}Train\\ Dataset\end{tabular}}} &
  \multicolumn{6}{c}{\textbf{Test Datasets}} \\ \cline{3-8} 
 &
   &
  \textit{\textbf{CASIA 1.0}} &
  \textit{\textbf{CASIA 2.0}} &
  \textit{\textbf{MICC-220}} &
  \textit{\textbf{MICC-2000}} &
  \textit{\textbf{COVERAGE}} &
  \textit{\textbf{Columbia Color}} \\ \hline
\multirow{3}{*}{Alahmadi et al. \cite{Alahmadi2017}} &
  \textit{\textbf{MICC-220}} &
  51.25\%  (0.34)* &
  60.66\%  (0.38)* &
  - &
  49.28\%  (0.33)* &
  53.33\%  (0.35)* &
  47.70\%  (0.32)* \\ \cline{2-2}
 &
  \textit{\textbf{COVERAGE}} &
  52.03\%  (0.49 &
  60.63\%  (0.39) &
  57.57\%  (0.58) &
  58.80\%  (0.59) &
  - &
  51.37\%  (0.51) \\ \cline{2-2}
 &
  \textit{\textbf{Columbia Color}} &
  52.24\%  (0.42) &
  37.17\%  (0.37) &
  45.45\%  (0.31) &
  57.61\%  (0.53) &
  51.59\%  (0.36) &
  - \\ \cline{1-2}
\multirow{3}{*}{Shilpa et al.  \cite{DUA2020369}} &
  \textit{\textbf{MICC-220}} &
  51.25\%  (0.34)* &
  60.65\%  (0.38)* &
  - &
  49.38\%  (0.33)* &
  54.00\%  (0.35)* &
  47.70\%  (0.32)* \\ \cline{2-2}
 &
  \textit{\textbf{COVERAGE}} &
  53.38\%  (0.39) &
  60.66\%  (0.38) &
  56.06\%  (0.53) &
  54.76\%  (0.51) &
  - &
  51.37\%  (0.47) \\ \cline{2-2}
 &
  \textit{\textbf{Columbia Color}} &
  52.03\%  (0.36) &
  64.75\%  (0.52) &
  46.96\%  (0.40) &
  47.14\%  (0.46) &
  51.82\%  (0.35) &
  - \\ \cline{1-2}
\multirow{3}{*}{Arman et al. \cite{9349611}} &
  \textit{\textbf{MICC-220}} &
  52.03\%  (0.44) &
  35.40\%  (0.30) &
  - &
  50.00\%  (0.45) &
  48.33\%  (0.48) &
  63.30\%  (0.63) \\ \cline{2-2}
 &
  \textit{\textbf{COVERAGE}} &
  51.06\%  (0.34)* &
  60.63\%  (0.38) &
  53.03\%  (0.47) &
  50.23\%  (0.37) &
  - &
  42.20\%  (0.33) \\ \cline{2-2}
 &
  \textit{\textbf{Columbia Color}} &
  53.57\%  (0.53) &
  38.44\%  (0.32) &
  39.39\%  (0.39) &
  57.14\%  (0.56) &
  47.94\%  (0.36) &
  - \\ \cline{1-2}
\multirow{3}{*}{Mandeep et al. \cite{10.1007/978-981-10-2738-3_27}} &
  \textit{\textbf{MICC-220}} &
  51.25\%  (0.34)* &
  60.66\%  (0.38)* &
  - &
  49.28\%  (0.33)* &
  53.33\%  (0.35)* &
  47.70\%  (0.32)* \\ \cline{2-2}
 &
  \textit{\textbf{COVERAGE}} &
  51.06\%  (0.34) &
  60.71\%  (0.39) &
  54.54\%  (0.48) &
  55.71\%  (0.50) &
  - &
  42.20\%  (0.34) \\ \cline{2-2}
 &
  \textit{\textbf{Columbia Color}} &
  44.29\% (0.44) &
  31.20\%  (0.29) &
  42.42\%  (0.42) &
  50.23\%  (0.49) &
  51.59\%  (0.38) &
  - \\ \cline{1-2}
\multirow{3}{*}{Mohammed et al. \cite{electronics9091500}} &
  \textit{\textbf{MICC-220}} &
  52.61\%  0.51) &
  42.82\%  (0.37) &
  - &
  45.23\%  (0.34) &
  53.53\%  (0.35)* &
  57.79\%  (0.52) \\ \cline{2-2}
 &
  \textit{\textbf{COVERAGE}} &
  50.87\%  (0.51) &
  57.75\%  (0.58) &
  42.42\%  (0.39) &
  35.23\%  (0.35 &
  - &
  33.02\%  (0.28) \\ \cline{2-2}
 &
  \textit{\textbf{Columbia Color}} &
  45.06\%  (0.45) &
  35.32\%  (0.35) &
  63.63\%  (0.61) &
  62.38\%  (0.62) &
  51.59\%  (0.34) &
  - \\ \cline{2-2}
\end{tabular}%
}
\caption{Training on Small Datasets and Testing on the rest}
\label{tab:smalltrain}
\end{table*}
\endgroup

%% file: tables/t_CommonTrainRealWorldTest.tex
\begingroup
\renewcommand{\arraystretch}{1.6}
\begin{table*}
\centering
\resizebox{\textwidth}{!}{%
\begin{tabular}{cccccccc}
\multirow{2}{*}{\textbf{Method}} &
  \multirow{2}{*}{\textbf{\begin{tabular}[c]{@{}c@{}}Test\\ Dataset\end{tabular}}} &
  \multicolumn{6}{c}{\textbf{Train Datasets}} \\ \cline{3-8} 
 &
   &
  \textit{\textbf{CASIA 1.0}} &
  \textit{\textbf{CASIA 2.0}} &
  \textit{\textbf{MICC-220}} &
  \textit{\textbf{MICC-2000}} &
  \textit{\textbf{COVERAGE}} &
  \textit{\textbf{Columbia Color}} \\ \hline
\multirow{4}{*}{Alahmadi et al. \cite{Alahmadi2017}} &
  \textit{\textbf{IMD-2020}} &
  52.80\%  (0.51) &
  74.40\%  (0.74) &
  44.40\%  (0.31)* &
  44.40\%  (0.31) &
  44.80\%  (0.34) &
  52.80\%  (0.49) \\ \cline{2-2}
 &
  \textit{\textbf{WildWeb}} &
  64.91\%  (0.61) &
  43.85\%  (0.44) &
  42.01\% (0.30)* &
  42.10\%  (0.30) &
  40.35\%  (0.29) &
  54.38\%  (0.49) \\ \cline{2-2}
 &
  \textit{\textbf{PS\_Battle}} &
  51.85\%  (0.50) &
  64.64\%  (0.63) &
  45.23\%  (0.34)* &
  45.11\%  (0.31)* &
  49.83\%  (0.43) &
  52.18\%  (0.50) \\ \cline{2-2}
 &
  \textit{\textbf{FRITH}} &
  \multicolumn{1}{l}{45.68\%  (0.44)} &
  \multicolumn{1}{l}{52.58\%  9.53)} &
  \multicolumn{1}{l}{40.41\%  (0.34)*} &
  \multicolumn{1}{l}{39.65\%  (0.28)*} &
  \multicolumn{1}{l}{55.17\%  (0.52)} &
  \multicolumn{1}{l}{44.82\%  (0.40)} \\ \cline{1-2}
\multirow{4}{*}{Shilpa et al.  \cite{DUA2020369}} &
  \textit{\textbf{IMD-2020}} &
  64.00\%  (0.64) &
  62.00\%  (0.62) &
  44.20\% (0.30) &
  44.40\%  (0.31)* &
  44.80\%  (0.31) &
  46.0\%  (0.40) \\ \cline{2-2}
 &
  \textit{\textbf{WildWeb}} &
  63.15\%  (0.58) &
  57.89\%  (0.51) &
  42.10\% (0.30) &
  42.00\%  (0.32)* &
  42.10\%  (0.30) &
  38.59\%  (0.32) \\ \cline{2-2}
 &
  \textit{\textbf{PS\_Battle}} &
  57.23\%  (0.57) &
  57.57\%  (0.50) &
  45.13\%  (0.31) &
  45.21\%  (0.31)* &
  45.11\%  (0.31) &
  46.80\%  (0.36) \\ \cline{2-2}
 &
  \textit{\textbf{FRITH}} &
  \multicolumn{1}{l}{48.27\%  (0.48} &
  \multicolumn{1}{l}{43.96\%  (0.44)} &
  39.0\%  (0.29) &
  \multicolumn{1}{l}{39.65\%  (0.28)*} &
  \multicolumn{1}{l}{42.24\%  (0.33)} &
  \multicolumn{1}{l}{43.96\%  (0.42)} \\ \cline{1-2}
\multirow{4}{*}{Arman et al. \cite{9349611}} &
  \textit{\textbf{IMD-2020}} &
  57.20\%  (0.42) &
  66.80\%  (0.66) &
  57.20\%  (0.42) &
  44.00\%  (0.32)* &
  44.40\%  (0.31)* &
  57.60\%  (0.41) \\ \cline{2-2}
 &
  \textit{\textbf{WildWeb}} &
  57.89\%  (0.40) &
  56.14\%  (0.50 &
  57.89\%  (0.40) &
  42.10\%  (0.30)* &
  42.10\%  (0.30)* &
  57.90\%  (0.40) \\ \cline{2-2}
 &
  \textit{\textbf{PS\_Battle}} &
  55.89\%  (0.38) &
  54.88\%  (0.43) &
  58.92\%  (0.47) &
  44.78\%  (0.31) &
  45.00\%  (0.31) &
  57.57\%  (0.42) \\ \cline{2-2}
 &
  \textit{\textbf{FRITH}} &
  \multicolumn{1}{l}{58.62\%  (0.37)} &
  \multicolumn{1}{l}{45.68\%  (0.46)} &
  56.03\%  (0.36) &
  39.60\%  (0.29)* &
  \multicolumn{1}{l}{40.51\%  (0.30)} &
  \multicolumn{1}{l}{59.48\%  (0.37)} \\ \cline{1-2}
\multirow{4}{*}{Mandeep et al. \cite{10.1007/978-981-10-2738-3_27}} &
  \textit{\textbf{IMD-2020}} &
  52.40\%  (0.52) &
  72.00\%  (0.71) &
  44.40\%  (0.31)* &
  44.60\% (0.31)* &
  44.00\%  (0.31) &
  53.20\%  (0.48) \\ \cline{2-2}
 &
  \textit{\textbf{WildWeb}} &
  57.89\%  (0.57) &
  49.12\%  (0.49 &
  42.00\%  (0.31)* &
  42.10\%  (0.30)* &
  42.20\%  (0.30)* &
  57.89\%  (0.53) \\ \cline{2-2}
 &
  \textit{\textbf{PS\_Battle}} &
  47.81\%  (0.47) &
  57.91\%  (0.56) &
  46.13\%  (0.33) * &
  45.00\%  (0.32)* &
  44.78\%  (0.31) &
  52.52\%  (0.51) \\ \cline{2-2}
 &
  \textit{\textbf{FRITH}} &
  \multicolumn{1}{l}{43.10\%  (0.35)} &
  44.82\%  (0.44) &
  \multicolumn{1}{l}{40.65\%  (0.28)*} &
  39.60\%  (0.28)* &
  \multicolumn{1}{l}{40.51\%  (0.30)} &
  \multicolumn{1}{l}{56.89\%  (0.54)} \\ \cline{1-2}
\multirow{4}{*}{Mohammed et al. \cite{electronics9091500}} &
  \textit{\textbf{IMD-2020}} &
  64.80\%  (0.64) &
  72.80\%  (0.73) &
  46.40\%  (0.46) &
  49.20\%  (0.48) &
  59.20\%  (0.56) &
  44.40\%  (0.44) \\ \cline{2-2}
 &
  \textit{\textbf{WildWeb}} &
  61.40\%  (0.55) &
  45.61\%  (0.46) &
  52.63\% (0.53) &
  49.12\%  (0.45) &
  50.87\%  (0.50) &
  49.12\%  (0.48) \\ \cline{2-2}
 &
  \textit{\textbf{PS\_Battle}} &
  55.55\%  (0.50) &
  62.28\%  (0.62) &
  49.15\%  (0.49) &
  47.47\%  (0.45 &
  49.49\%  (0.48) &
  52.52\%  (0.52) \\ \cline{2-2}
 &
  \textit{\textbf{FRITH}} &
  \multicolumn{1}{l}{54.31\%  (0.50)} &
  \multicolumn{1}{l}{48.27\%  (0.48)} &
  51.72\%  (0.36) &
  45.68\%  (0.46) &
  47.41\%  (0.47) &
  52.58\%  (0.52)
\end{tabular}%
}
\caption{Training on Commonly used Image Forgery Datasets and Testing on Real-World Manipulated Data}
\label{tab:realtest}
\end{table*}
\endgroup

%% file: tables/t_MixtureRealDatasets.tex
\begingroup
\renewcommand{\arraystretch}{1.7}
\begin{table}
\centering
\resizebox{\columnwidth}{!}{%
\begin{tabular}{ccccc}
\multirow{2}{*}{\textbf{Methods}} & \multicolumn{4}{c}{\textbf{Data Samples}}                             \\ \cline{2-5} 
 & \textit{\textbf{Random Sample 1}} & \textit{\textbf{Random Sample 2}} & \textit{\textbf{Random Sample 3}} & \textit{\textbf{Mean Performance}} \\ \hline
Alahmadi et al. \cite{Alahmadi2017}                   & 67.50\%  (0.67) & 60.00\%  (0.60) & 63.33\%  (0.63) & 63.61\%  (0.63) \\ \cline{1-1}
Shilpa et al. \cite{DUA2020369}                    & 45.83\%  (0.46) & 55.83\%  (0.56) & 57.50\%  (0.57) & 53.05\%  (0.53) \\ \cline{1-1}
Arman et al. \cite{9349611}                      & 56.67\% (0.57)  & 54.16\%  (0.54) & 57.50\%  (0.57) & 56.11\%  (0.56) \\ \cline{1-1}
Mandeep et al. \cite{10.1007/978-981-10-2738-3_27}                   & 58.33\%  (0.58) & 56.67\%  (0.57) & 58.33\%  (0.56) & 57.77\%  (0.57) \\ \cline{1-1}
Mohammed et al. \cite{electronics9091500}                   & 60.00\%  (0.60) & 53.33\%  (0.53) & 60.83\%  (0.61) & 58.05\%  (0.58)
\end{tabular}%
}
\caption{Amalgamation of Real-world Datasets}
\label{tab:mixreal}
\end{table}
\endgroup